\documentclass[conference]{IEEEtran}
\IEEEoverridecommandlockouts
\usepackage{amsmath,amssymb,amsfonts}
\usepackage{algorithmic}
\usepackage{graphicx}
\usepackage{textcomp}
\usepackage{xcolor}
\date{\today}

\usepackage{biblatex}
\usepackage{amsmath}  
\usepackage{amssymb}  
\usepackage{graphicx}  
\usepackage{subcaption}
\usepackage{multirow}
\usepackage{graphicx} 
\usepackage{array}
\def\BibTeX{{\rm B\kern-.05em{\sc i\kern-.025em b}\kern-.08em
    T\kern-.1667em\lower.7ex\hbox{E}\kern-.125emX}}
\begin{document}

\title{Dimensionality Reduction for Hyperspectral Image Classification\\

}

\author{
  \IEEEauthorblockN{Mohamed Cherifi}
  \IEEEauthorblockA{\textit{Laboratoire Traitement du Signal} \\
    \textit{Ecole Militaire Polytechnique}\\
    Bordj-El-Bahri, Alger, 16111, Algérie \\
    }
  \and
  \IEEEauthorblockN{Ammar Mesloub}
  \IEEEauthorblockA{\textit{Laboratoire Antennes et Dispositifs Micro-Ondes} \\
    \textit{Ecole Militaire Polytechnique}\\
    Bordj-El-Bahri, Alger, 16111, Algérie \\
   }
  \and
  \IEEEauthorblockN{Mohammed Nabil El Korso}
  \textit{Université Paris-Saclay}\\
  Saclay, France \\
 
  \and
  \IEEEauthorblockN{Tayeb Touhami}
  \IEEEauthorblockA{\textit{Laboratoire Antennes et Dispositifs Micro-Ondes} \\
    \textit{Ecole Militaire Polytechnique}\\
    Bordj-El-Bahri, Alger, 16111, Algérie \\
  }
  \and
  \IEEEauthorblockN{Abdennour Hacine Gharbi}
  \IEEEauthorblockA{\textit{Université de Bordj-Bou-Ariridj}\\
    Bordj-Bou-Ariridj, Algérie \\
  }
}

\maketitle

\begin{abstract}
This paper addresses the issue of supervised classification in the context of hyperspectral satellite images. It deals with two fundamental aspects: dimensionality reduction of data and the selection of appropriate supervised classification techniques.

Firstly, we delve into dimensionality reduction, a critical step in simplifying the management of hyperspectral data. The reduction aims to decrease complexity in terms of memory and computing time. We examine two commonly used methods: Principal Component Analysis (PCA) and Linear Discriminant Analysis (LDA).

Subsequently, we explore the selection of the most suitable supervised classification algorithms for hyperspectral images. We compare the performance of three methods: K-Nearest Neighbors (KNN), Support Vector Machines (SVM), and Random Forest (RF) using real hyperspectral data. The results highlight that the combination of PCA and RF yields the highest overall accuracy and Kappa coefficient.
\end{abstract}

\begin{IEEEkeywords}
    PCA (Principal Component Analysis),
    LDA (Linear Discriminant Analysis),
    KNN (k-Nearest Neighbors),
    RF (Random Forest),
    SVM (Support Vector Machine)
\end{IEEEkeywords}

\section{Introduction}
Remote Sensing is a technique introduced in the early 1960s for data analysis and interpretation \cite{holland1995bunce}. It has revolutionized our ability to collect vast amounts of satellite data, providing extensive geographical coverage with high temporal frequency compared to other imaging methods. The interpretation of satellite images serves a multitude of purposes, ranging from environmental conservation and management, water resource research, and soil quality studies to post-natural disaster environmental assessments, meteorology simulations, land use and land cover analysis, disaster prevention, and the study of climatic changes\cite{abburu2015satellite}.

Within the realm of remote sensing, hyperspectral remote sensors take center stage. These sensors, known for their high spectral resolution, are instrumental in monitoring the Earth's surface. Hyperspectral images (HSI), characterized by their inclusion of more than three bands compared to conventional RGB images, are indispensable tools. They find application across diverse domains, including crop analysis, geological mapping, mineral exploration, defense research, urban investigation, military surveillance.

The interest of dimensionality reduction in the context of hyperspectral images lies in the efficient management of the information contained in these complex images. It not only reduces complexity in terms of memory and computing time but also enhances the performance of analysis and classification techniques. Furthermore, it helps eliminate unwanted noise or redundancies in hyperspectral data, thereby improving classification accuracy and the quality of image interpretation.

In the classification realm, methods are traditionally categorized as supervised or unsupervised. Unsupervised classification \cite{mather2016classification}, often referred to as clustering, seeks to group data in the absence of sample sets. In contrast, supervised classification relies on analyst input and employs sample training sets to identify different classes. In most applications, supervised classification offers distinct advantages over unsupervised approaches. The MATLAB toolbox for supervised classification encompasses a wide array of classical classification algorithms and various classifying methods, all of which are evaluated in this paper, providing a comprehensive comparative analysis of their accuracy and suitability for remote sensing applications for hyperspectral image dataset \cite{rees2013physical} .

\section{Dataset}
In this study, we utilize the Indian Pines (IP) hyperspectral image dataset\cite{baumgardner2015220}, which has been widely employed in hyperspectral image analysis. The IP dataset was originally gathered using the AVIRIS sensor over the Indian Pines test site located in north-western Indiana. It comprises a wealth of information with the following key characteristics:

\begin{itemize}
  \item Dimensions: 145 x 145 pixels
  \item Spectral Bands: 220 bands
  \item Classes: 16 distinct classes
\end{itemize}

This dataset serves as an essential resource for various hyperspectral image processing tasks, including classification, feature extraction, and dimensionality reduction. The richness of spectral information and the diversity of land cover classes make it a valuable benchmark for evaluating classification algorithms and conducting experiments in the field of remote sensing.
\subsection{Class Reduction Approach }

In this section, we will explain our approach to class reduction for hyperspectral image classification using the  (IP) dataset. Our methodology is based on the physical characteristics of spectral reflections at different wavelengths. The original classes (16 in total) are consolidated into 6 classes that share similar spectral responses. This reduction simplifies the classification, making it more efficient and interpretable. Table 1 provides a brief description of each class in the original dataset, aiding in the understanding of the dataset's content and potential applications. 
Additionally, as per your request, we have included Table 2, which shows the class descriptions and the sum of samples for the reduced classes.
\begin{table}[t]
  \centering
  \small 
  \begin{tabular}{|c|c|c|}
    \hline
    \textbf{Class} & \textbf{Description} & \textbf{Samples} \\
    \hline
    1 & Alfalfa & 46 \\
    2 & Corn-notill & 428 \\
    3 & Corn-min & 830 \\
    4 & Corn & 237 \\
    5 & Grass/Pasture & 483 \\
    6 & Grass/Trees & 730 \\
    7 & Grass/pasture-mowed & 28 \\
    8 & Hay-windrowed & 478 \\
    9 & Oats & 20 \\
    10 & Soybeans-notill & 972 \\
    11 & Soybeans-min & 2455 \\
    12 & Soybean-clean & 593 \\
    13 & Wheat & 205 \\
    14 & Woods & 1265 \\
    15 & Bldg-Grass-Tree-Drives & 386 \\
    16 & Stone-steel towers & 93 \\
    \hline
  \end{tabular}
  \caption{Class Descriptions and Sample Counts}
  \label{tab:class-samples}
\end{table}

\begin{table*}[t] 
  \centering
  \small 
  \begin{tabular}{|c|c|c|}
    \hline
    \textbf{Class} & \textbf{Description} & \textbf{Samples (Sum of Parent Classes)} \\
    \hline
    1 & Alfalfa & 46 \\
    2 & Corn-notill-Wheat  & 428 + 830 + 237 + 205 = 1700 \\
    3 & Grass-Pasture & 483 + 730 + 28 = 1241 \\
    4 & Hay-windrowed & 478 \\
    5 & Soybeans-notill-Oats & 972 + 2455 + 593 + 20 = 4020 \\
    6 & Woods-Bldg-Grass-Stone  &   1265 + 386+ 93 = 1744\\
    \hline
  \end{tabular}
  \caption{Ground Truth Details for the Reduced Indian Pines (IP) Dataset}
  \label{tab:reduced_classes}
\end{table*}

\section{Dimensionality Reduction}

Dimensionality Reduction addresses the challenges associated with analyzing multivariate data. When working with extensive datasets, it becomes necessary to reduce dimensionality. The primary objective of dimensionality reduction is to represent the data in a lower-dimensional space while preserving some of its essential properties. Equation 1 illustrates the reduction of a high-dimensional dataset \(\mathbf{n}\) into a lower-dimensional space \(\mathbf{m}\).

\begin{equation}
\mathbf{X} = \begin{bmatrix}
x_1 \\
x_2 \\
\vdots \\
x_n
\end{bmatrix}
\quad \text{reduce dimensionality} \quad
\mathbf{Y} = \begin{bmatrix}
y_1 \\
y_2 \\
\vdots \\
y_m
\end{bmatrix}
\end{equation}
where
\[
n > m
\]

Dimensionality reduction plays a crucial role at the intersection of various fields, including data mining, databases, statistics, pattern recognition, text mining, visualization, artificial intelligence, and optimization. It serves as an essential technique to manage and analyze complex data. Dimensionality reduction encompasses several approaches, including supervised methods like Linear Discriminant Analysis (LDA), Support Vector Machines (SVM), and Hopfield Neural Networks (HNN), as well as unsupervised techniques such as Principal Component Analysis (PCA), Singular Value Decomposition (SVD), and Independent Component Analysis (ICA).
In this paper, we focus on unsupervised PCA and supervised LDA for dimensionality reduction in the context of image dataset\cite{sachin2015dimensionality}.

\subsection{Principal Component Analysis }

 PCA stands as a widely recognized technique for dimensionality reduction, revered in a multitude of fields. Its primary objective is to diminish the dimensionality of a dataset while conserving the maximum possible variance \cite{pang2008effective}. The essence of PCA lies in its identification of orthogonal axes, termed principal components, that best encapsulate the dataset's variance characteristics.

These principal components are ordered, with the first capturing the most substantial variance, followed by the second, and so on. By judiciously selecting a subset of these principal components \cite{smith2002tutorial}, one can effectively represent the data in a lower-dimensional space.

 a. For the covariance matrix:
\begin{equation}
\boldsymbol{\Sigma} = \frac{1}{n} \sum_{i=1}^{n} (\mathbf{X}_i - \bar{\mathbf{X}})(\mathbf{X}_i - \bar{\mathbf{X}})^T
\end{equation}

b. For the eigenvalue decomposition (eigenvalues and eigenvectors):
\begin{equation}
\boldsymbol{\Sigma} \mathbf{v} = \boldsymbol{\lambda} \mathbf{v}
\end{equation}

\begin{itemize}
    \item Where $\boldsymbol{\Sigma}$ is the covariance matrix.
    \item $\mathbf{X}_i$ represents individual data points.
    \item $\bar{\mathbf{X}}$ is the mean (average) of the data points.
    \item $\mathbf{v}$ is an eigenvector.
    \item $\boldsymbol{\lambda}$ is the corresponding eigenvalue.
\end{itemize}

    c. For projecting data onto the principal components:
\begin{equation}
\mathbf{Y} = \mathbf{X}\mathbf{W}
\end{equation}

\begin{itemize}
    \item where $\mathbf{Y}$ is the matrix of projected data.
    \item $\mathbf{X}$ is the original data matrix.
    \item $\mathbf{W}$ is the matrix of eigenvectors chosen as principal components.
\end{itemize}

PCA has demonstrated its efficacy in diverse domains, spanning image processing, face recognition, and data compression. It proves especially indispensable when confronting high-dimensional datasets, offering efficient means for data representation and visualization.
Figure 2 illustrates the visualization of the first three principal components obtained from Principal Component Analysis (PCA). These principal components are essential for reducing the dimensionality of hyperspectral data and enable an efficient representation of the information contained within the image. Figure 1 demonstrates how these principal components capture significant data variance, thereby contributing to the understanding of essential features within the hyperspectral image.
\subsection{Linear Discriminant Analysis (LDA)}

In contrast to PCA, (LDA) is a dimensionality reduction technique tailored for supervised classification. LDA departs from unsupervised approaches by incorporating class labels in its quest to unearth a lower-dimensional space that maximizes the separation between distinct classes.

LDA's core objective revolves around discovering linear combinations of features, aptly referred to as discriminants, that foster clear demarcation between class clusters, all while minimizing within-class variance \cite{bishop2006pattern}. As a result, LDA finds its niche in applications such as pattern recognition, face verification, and medical diagnosis.

The field of remote sensing has extensively explored LDA, leveraging its capabilities for tasks like land cover classification, hyperspectral image analysis, and target detection. Its unique ability to enhance class separability renders LDA an invaluable tool within the realm of remote sensing.

  a. To calculate the between-class scatter matrix (\(S_B\)):

\begin{equation*}
\mathbf{S_B} = \sum_{i=1}^{c} N_i (\mathbf{m}_i - \mathbf{m})(\mathbf{m}_i - \mathbf{m})^T \tag{5}
\end{equation*}

\begin{itemize}
    \item where \(c\) is the number of classes.
    \item \(N_i\) is the number of samples in class \(i\).
    \item \(\mathbf{m}_i\) is the mean of samples in class \(i\).
    \item \(\mathbf{m}\) is the overall mean of all data.
\end{itemize}

b. To calculate the within-class scatter matrix (\(S_W\)):

\begin{equation*}
\mathbf{S_W} = \sum_{i=1}^{c} \sum_{j=1}^{N_i} (\mathbf{x}_{ij} - \mathbf{m}_i)(\mathbf{x}_{ij} - \mathbf{m}_i)^T \tag{6}
\end{equation*}

\begin{itemize}
    \item where \(\mathbf{x}_{ij}\) is the \(j\)-th sample in class \(i\).
\end{itemize}

c. To calculate the generalized inverse of the within-class scatter matrix times the between-class scatter matrix (\(S_W^{-1}\mathbf{S_B}\)):

\begin{equation*}
\mathbf{S_W}^{-1}\mathbf{S_B} \tag{7}
\end{equation*}

d. To obtain the eigenvectors (\(\mathbf{v}_i\)) and eigenvalues (\(\lambda_i\)) of \(\mathbf{S_W}^{-1}\mathbf{S_B}\):

\begin{equation*}
\mathbf{S_W}^{-1}\mathbf{S_B} \mathbf{v}_i = \lambda_i \mathbf{v}_i \tag{8}
\end{equation*}

\begin{itemize}
    \item where \(\mathbf{v}_i\) is an eigenvector.
    \item \(\lambda_i\) is the corresponding eigenvalue.
\end{itemize}

e. To project the data onto the linear discriminant components (\(\mathbf{y}\)):

\begin{equation*}
\mathbf{y} = \mathbf{X} \mathbf{V} \tag{9}
\end{equation*}

\begin{itemize}
    \item where \(\mathbf{y}\) is the matrix of projected data.
    \item \(\mathbf{X}\) is the original data matrix.
    \item \(\mathbf{V}\) is the matrix of eigenvectors chosen as linear discriminant components.
\end{itemize}
The Figure 3 presents the visualization of the first three principal components obtained from Linear Discriminant Analysis (LDA). These principal components are crucial for reducing the dimensionality of hyperspectral data and aiding in the differentiation between various classes
\begin{figure*}[ht!] 
    \centering
    \begin{subfigure}{0.3\textwidth}
        \includegraphics[width=\linewidth]{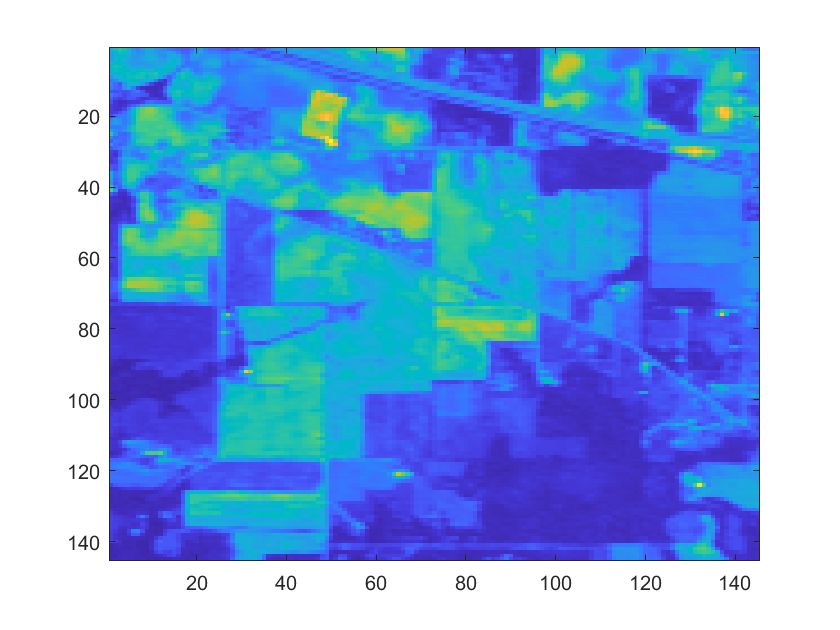}
        \caption{Band 23}
        \label{fig:bande23}
    \end{subfigure}\hfill
    \begin{subfigure}{0.3\textwidth}
        \includegraphics[width=\linewidth]{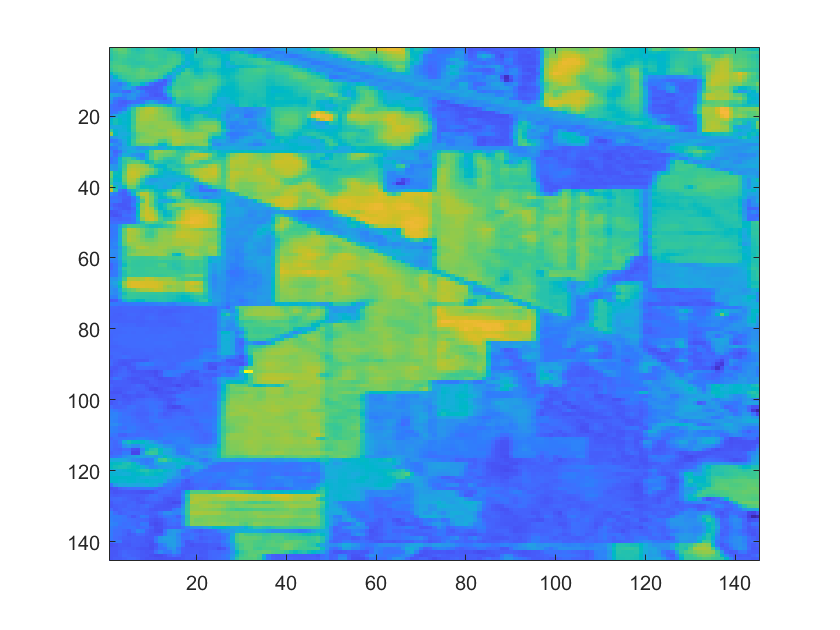}
        \caption{Band 124}
        \label{fig:bande124}
    \end{subfigure}\hfill
    \begin{subfigure}{0.3\textwidth}
        \includegraphics[width=\linewidth]{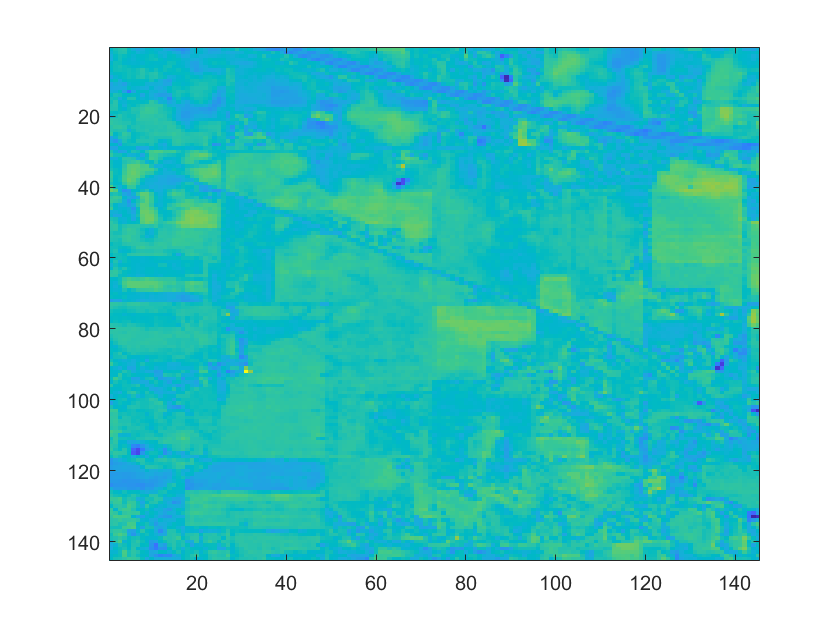}
        \caption{Bande 95}
        \label{fig:bande95}
    \end{subfigure}
    
    \caption{The visualization of the three randomly selected bands over 220.}
    \label{fig:visualisation_bandes}
\end{figure*}

\begin{figure*}[ht] 
    \centering
    \begin{subfigure}{0.3\textwidth}
        \includegraphics[width=\linewidth]{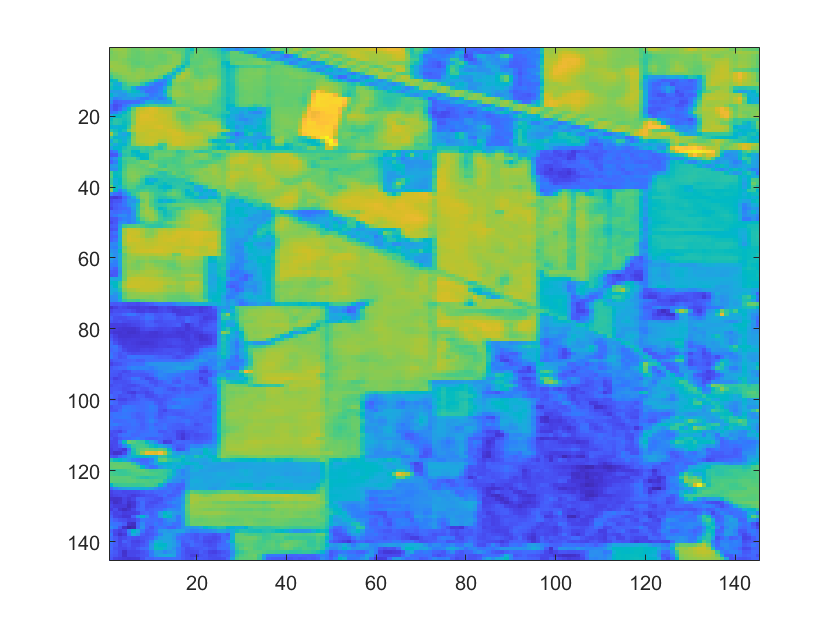}
        \caption{Band 1}
        \label{fig:pca1}
    \end{subfigure}\hfill
    \begin{subfigure}{0.3\textwidth}
        \includegraphics[width=\linewidth]{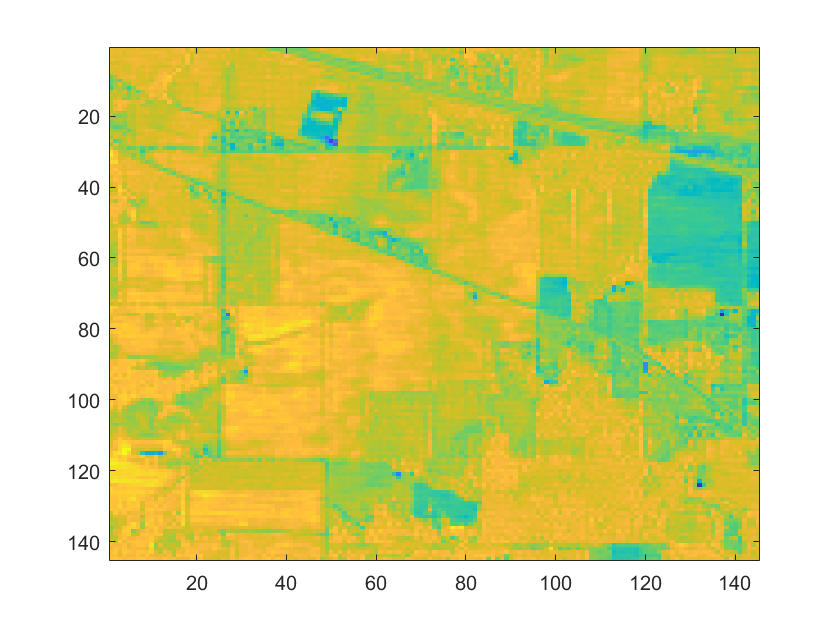}
        \caption{Band 2}
        \label{fig:pca4}
    \end{subfigure}\hfill
    \begin{subfigure}{0.3\textwidth}
        \includegraphics[width=\linewidth]{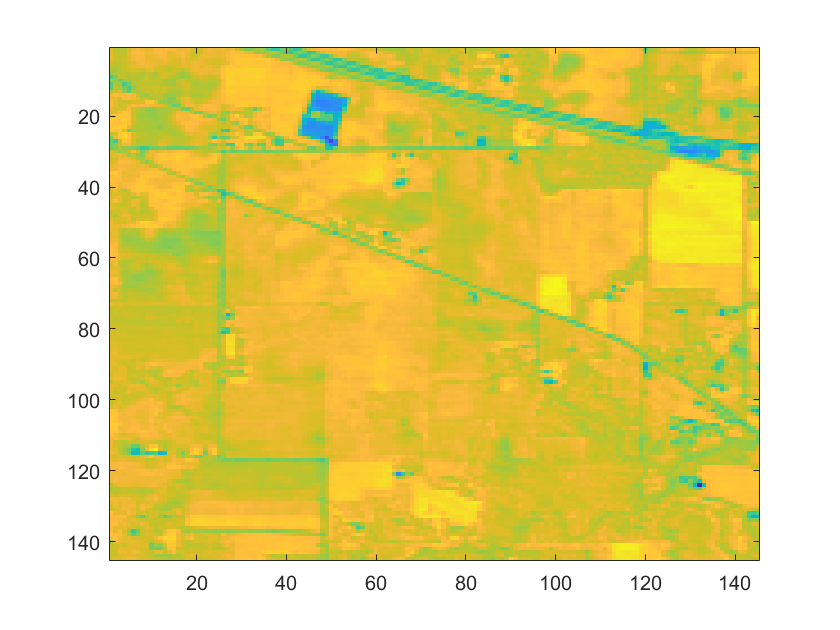}
        \caption{Band 3}
        \label{fig:pca3}
    \end{subfigure}
    
    \caption{Visualization of the selected bands after PCA.}
    \label{fig:pca_bands}
\end{figure*}

\begin{figure*}[ht] 
    \centering
    \begin{subfigure}{0.3\textwidth}
        \includegraphics[width=\linewidth]{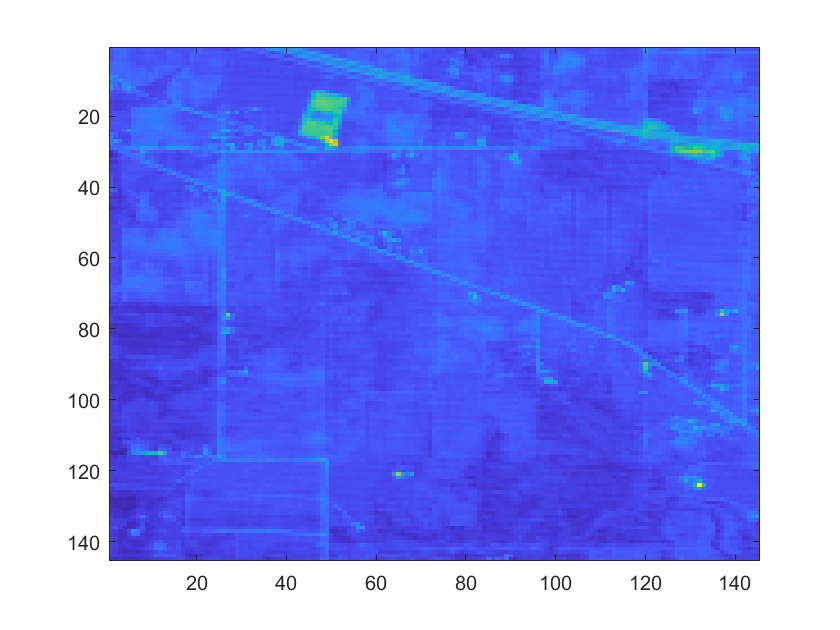}
        \caption{Band 1}
        \label{fig:lda1}
    \end{subfigure}\hfill
    \begin{subfigure}{0.3\textwidth}
        \includegraphics[width=\linewidth]{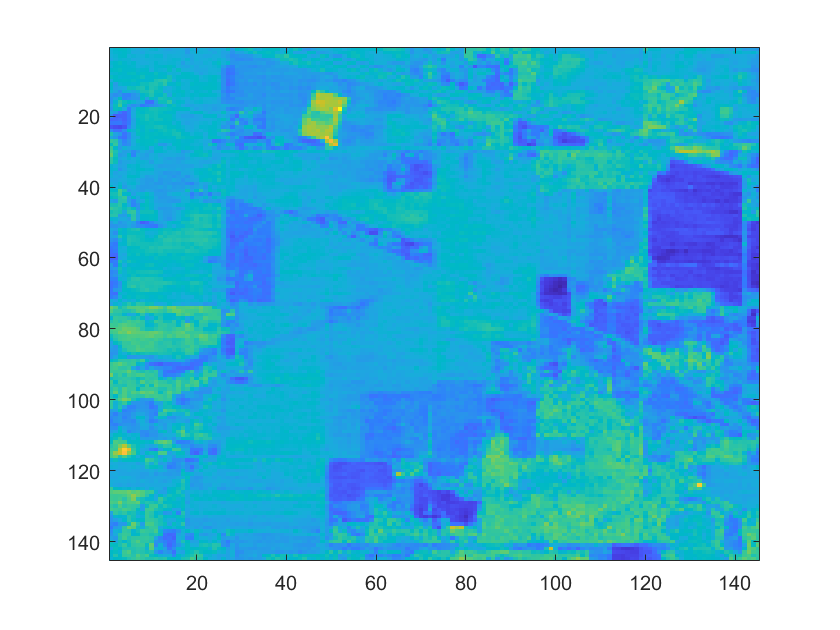}
        \caption{Band 2}
        \label{fig:lda2}
    \end{subfigure}\hfill
    \begin{subfigure}{0.3\textwidth}
        \includegraphics[width=\linewidth]{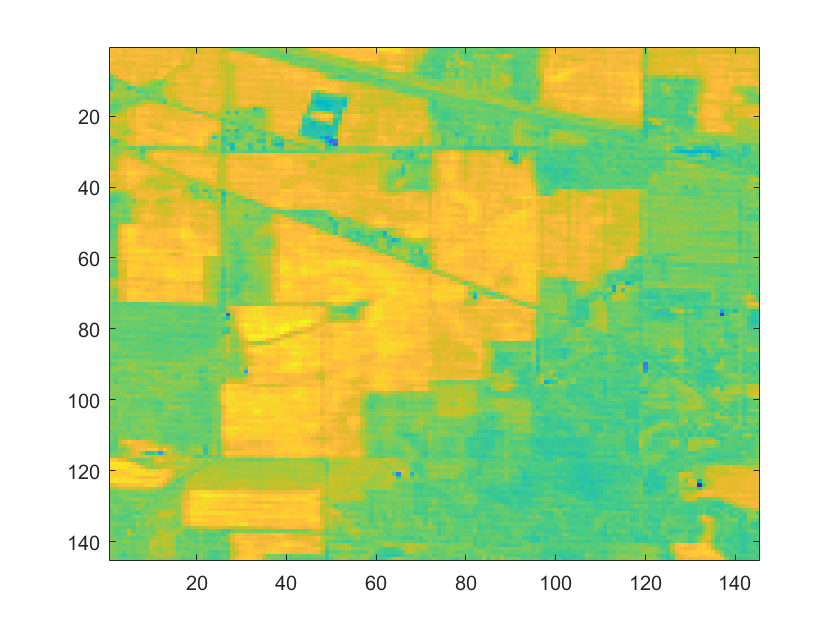}
        \caption{Band 3}
        \label{fig:lda3}
    \end{subfigure}
    
    \caption{Visualization of the selected bands after LDA.}
    \label{fig:lda_bands}
\end{figure*}

\section{METHODOLOGY}

\subsection{K-Nearest Neighbors (KNN)}

The KNN algorithm is a non-parametric method commonly used for classification, relying on the proximity of training examples in the feature space. The KNN classification process involves partitioning data into a test set and a training set. For each row in the test set, the K nearest training set instances are determined using the Euclidean distance metric, and the classification is decided through a majority vote. In cases where there is a tie for the Kth nearest neighbor, all tied candidates are included in the voting process. A noteworthy characteristic of KNN is its reliance on the entire training dataset during the testing phase, where decisions are made based on the entirety of the training data \cite{hastie2009elements}.

The mathematical expression of the  (KNN) algorithm can be formulated as follows:

a. Let \(X\) be the training dataset with features \(\mathbf{x}_i\) and class labels \(y_i\) for \(i = 1, 2, \ldots, N\), where \(N\) is the number of training examples.

b. Let \(x\) be a test example that we want to classify.

c. Calculate the Euclidean distance between \(x\) and each training example \(\mathbf{x}_i\):

\[
\text{distance}(\mathbf{x}, \mathbf{x}_i) = \sqrt{\sum_{j=1}^{d} (x_j - \mathbf{x}_{ij})^2}\tag{10}
\]

   - Where \(d\) is the number of features.

d. Select the \(K\) training examples with the shortest distances to \(x\).

e. Perform a majority vote among the class labels of these \(K\) neighbors.

   - The most frequent class among the \(K\) neighbors is assigned to the test example \(x\) as its predicted class. In case of a tie, all classes are considered.

 The choice of the parameter \(K\) and the distance metric used (such as Euclidean distance) are important considerations when applying KNN to classification tasks.

\subsection{Support Vector Machines (SVM)}

  SVM is a parametric classification approach used to tackle classification challenges in datasets where the relationships between variables are not explicitly known.

SVM is rooted in statistical learning theory and was initially designed to classify linearly separable data with two classes. However, it has since been extended to handle nonlinear and multi-class datasets effectively. The core principle of SVM involves identifying the hyperplane that optimally discriminates between the two classe\cite{hao2010partial}.

In our specific case, we employed the Gaussian kernel function, also known as the radial basis function (RBF) kernel. This choice of kernel allows SVM to effectively capture complex, nonlinear relationships within hyperspectral data.

The mathematical expression for the RBF kernel function is given by:
\begin{equation*}
K(x, x') = \exp\left(-\frac{\|x - x'\|^2}{2\sigma^2}\right) \tag{10}
\end{equation*}
  
\begin{itemize}
    \item \(K(x, x')\) is the RBF kernel function.
    \item \(x\) and \(x'\) are the input data points.
    \item \(\sigma\) is a parameter that controls the width of the kernel and influences the flexibility of the decision boundary.
\end{itemize}

\begin{figure*}[ht] 
    \centering
    \includegraphics[width=0.5\textwidth]{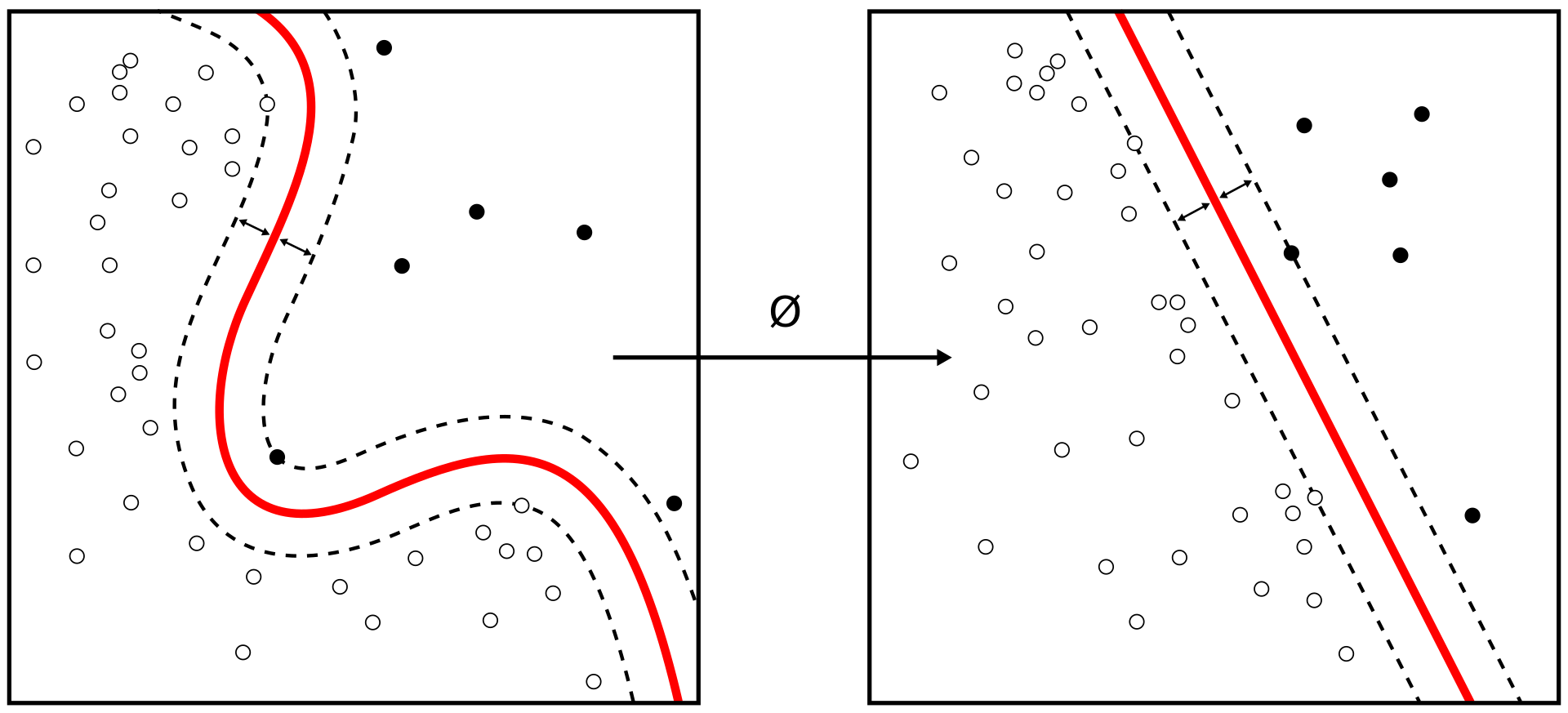} 
    \caption{Illustration of Support Vector Machine}
    \label{fig:votre_image}
\end{figure*}

To accommodate nonlinear data relationships, SVM employs kernel functions that map the data into a higher-dimensional space\cite{kavzoglu2020comparison}. The optimization process in SVM aims to maximize the margins between support vectors and establish an optimal decision function based on the hyperplane in the transformed feature space\cite{hastie2009elements}.

This choice of the RBF kernel and the optimization process in SVM enable it to perform effective classification even in cases with complex, nonlinear data relationships.

\subsection{Random Forest (RF)}

We delve into the methodology of  (RF) for hyperspectral image classification. RF is a powerful ensemble learning technique built upon the principle of employing decision trees as elementary classifiers and aggregating their results to create a robust collective learning model \cite{breiman2001random}. RF classifiers are renowned for their resilience against overfitting, ease of parameterization, and computational efficiency (Kavzoglu, 2017).

The primary objective of the RF classifier is to construct a multitude of decision trees using a bootstrapped sampling approach. During this process, the training dataset used for creating tree models within the decision forest is selected randomly from the original training dataset. Roughly two-thirds of the randomly sampled dataset are utilized for constructing the decision tree structure, while the remaining portion is reserved for validating the generated decision tree models. To classify an uncertain sample, the class label is determined using the majority voting principle, where each tree model in the decision forest contributes its prediction.
\begin{table*}[t] 
    \captionsetup{justification=centering} 
    \centering
    \caption{Classification Overall Accuracy and Kappa Values with Dimensionality Reduction (OA: Overall Accuracy, Ka: Kappa)}
    \begin{tabular}{|c|c|c|c|c|}
        \hline
        \multirow{2}{*}{\textbf{Reduction Method}} & \multirow{2}{*}{\textbf{Classifier}} & \textbf{OA (\%)} & \textbf{Ka (\%)} \\
        & & & \\
        \hline
        \multirow{3}{*}{PCA} & KNN & 0.941 & 0.904 \\
        & RF & 0.955 & 0.926 \\
        & SVM & 0.938 & 0.896 \\
        \hline
        \multirow{3}{*}{LDA} & KNN & 0.938 & 0.898 \\
        & RF & 0.945 & 0.938 \\
        & SVM & 0.926  & 0.905 \\
        \hline
    \end{tabular}
\end{table*}

\begin{table*}[t] 
    \captionsetup{justification=centering} 
    \centering
    \caption{Classification overall accuracies and F-score values of datasets with PCA and LDA}
    \begin{tabular}{|c|c|c|c|c|c|c|}
        \hline
        \multirow{2}{*}{\textbf{LULC Classes}} & \multicolumn{2}{c|}{\textbf{SVM}} & \multicolumn{2}{c|}{\textbf{RF}} & \multicolumn{2}{c|}{\textbf{KNN}} \\
        \cline{2-7}
         & \textbf{PCA} & \textbf{LDA} & \textbf{PCA} & \textbf{LDA} & \textbf{PCA} & \textbf{LDA} \\
        \hline
        Class 1 & 0.949 & 0.941 & 0.963 & 0.96 & 0.951 & 0.950 \\
        \hline
        Class 2 & 0.912 & 0.899 & 0.944 & 0.91 & 0.921 & 0.913 \\
        \hline
        Class 3 & 0.937 & 0.918 & 0.959 & 0.951 & 0.949 & 0.934 \\
        \hline
        Class 4 & 0.939 & 0.929 & 0.960 & 0.943 & 0.946 & 0.945 \\
        \hline
        Class 5 &0.875 & 0.850 & 0.888    & 0.88 & 0.863 & 0.858 \\
        \hline
        Class 6 & 0.901 & 0.848 & 0.952 & 0.962 & 0.914 & 0.941 \\
        \hline
        OA (\%) & 0.938 & 0.926 & 0.955 & 0.945 & 0.941 & 0.938 \\
        \hline
    \end{tabular}
\end{table*}

\begin{figure*}[htbp] 
    \centering
    \includegraphics[width=0.7\textwidth]{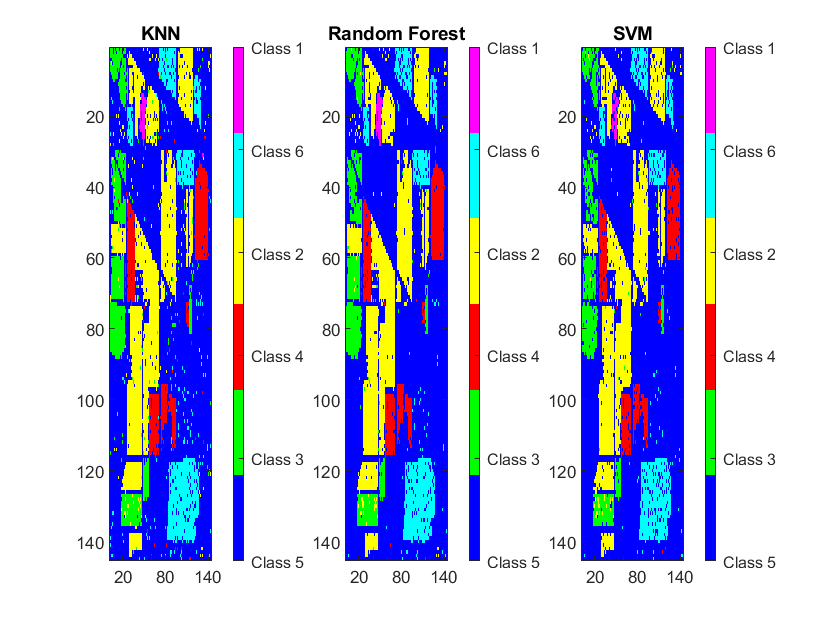}
    \caption{Classification after PCA Dimensionality Reduction with Various Classifiers}
    \label{fig:Figure 4}
\end{figure*}

\begin{figure*}[htbp] 
    \centering
    \includegraphics[width=0.7\textwidth]{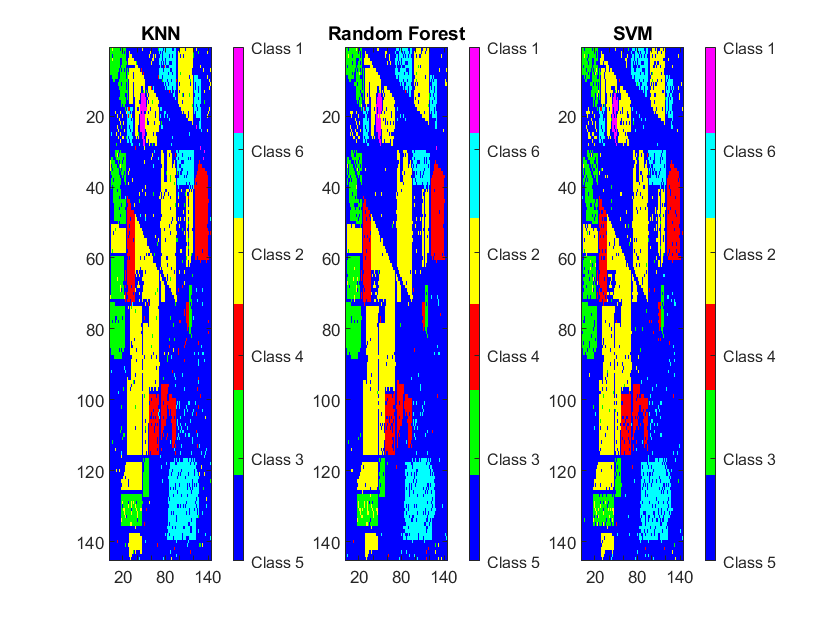}
    \caption{Classification after LDA Dimensionality Reduction with Various Classifiers}
    \label{fig:Figure 5}
\end{figure*}

The mathematical expression of the RF algorithm can be described as follows:

Let \(N\) be the number of decision trees in the forest, \(D\) be the training dataset, and \(x\) be an uncertain sample to be classified.

For each decision tree \(t_i\) in the forest \(i = 1, 2, \ldots, N\):

a. Randomly select a bootstrapped dataset \(D_i\) from \(D\) with replacement.

b. Train \(t_i\) on \(D_i\).

To classify \(x\):

c. Aggregate predictions from all decision trees:

\begin{equation*}
\hat{y}_i = t_i(x) \quad \text{for } i = 1, 2, \ldots, N   \tag{11}
\end{equation*}

d. Determine the final class label for \(x\) through majority voting:

\begin{equation*}
\hat{y} = \text{argmax}_y \sum_{i=1}^{N} I(\hat{y}_i = y) \tag{12}
\end{equation*}

Where:
\begin{itemize}
  \item\(N\) is the number of decision trees in the forest.
  \item\(D\) is the training dataset.
  \item\(x\) is the uncertain sample to be classified.
  \item\(t_i\) represents an individual decision tree.
  \item\(D_i\) is the bootstrapped dataset for tree \(t_i\).
  \item\(\hat{y}_i\) is the prediction of tree \(t_i\) for \(x\).
  \item\(\hat{y}\) is the final class label for \(x\).
  \item\(y\) represents class labels.
  \item\(I(\cdot)\) is the indicator function.
\end{itemize}

\subsection{Model Parameter Tuning (K-Fold Cross-Validation)}

In our study, we employed cross-validation techniques to meticulously tune the parameters of our hyperspectral image analysis models. This approach helps mitigate the risk of overfitting, especially in datasets with size constraints.

For the RF model, we focused on optimizing key parameters such as the number of decision trees in the forest, the maximum tree depth, and the minimum samples required to split a node. These parameters play a pivotal role in determining the model's complexity and its susceptibility to overfitting \cite{breiman2001random}. We systematically explored a range of values for each parameter to identify optimal values that would yield superior classification performance.

In the case of the KNN model, we fine-tuned the number of neighbors (K) to consider during classification. The choice of K directly influences the model's flexibility and sensitivity to noise. Therefore, we conducted experiments by adjusting K to find the value that offers the best generalization capacity.

For the SVM model with the Gaussian kernel function, we scrutinized two crucial parameters: the regularization coefficient (C) and the kernel width. The C parameter controls the tolerance to classification errors, while the kernel width affects the model's flexibility. Through careful parameter tuning, we aimed to discover the optimal combination that would yield superior classification performance.

To assess and compare the performance of each model at each stage of the parameter tuning process, we employed K-fold cross-validation with K set to 5 . Performance measures such as the mean R-squared scores were used as evaluation criteria\cite{jung2020evaluation}.

\subsubsection{Hyperparameters}

For our parameter tuning process, we considered the following hyperparameters for each classification model:

\textbf{KNN:}
\begin{itemize}
    \item Number of neighbors (\texttt{K})
\end{itemize}

\textbf{RF:}
\begin{itemize}
    \item Number of decision trees in the forest (\texttt{numTrees})
   
    \item Minimum samples required to split a node (\texttt{MinLeafSize})
\end{itemize}

\textbf{SVM with Gaussian Kernel:}
\begin{itemize}
    \item Regularization coefficient (\texttt{C})
  
\end{itemize}

These hyperparameters were systematically adjusted and optimized during the K-fold cross-validation process to enhance the classification performance of our models\cite{jung2020evaluation}.

\section{Application,Results and Discussion}

In this section, we present the results of our analysis. We began by selecting the top 10 features for both PCA and LDA to reduce dimensionality. 
After applying $k$-fold cross-validation, where $k=5$, we determined the optimal hyperparameter settings for the three classifiers that maximized the \textit{mean\_accuracy}. Specifically, for the K-Nearest Neighbors (KNN) classifier, the optimal value of $K$ was found to be 1. In the case of the Random Forest classifier, we identified the best hyperparameters as $\textit{numTrees}=150$ and $\textit{MinLeafSize}=1$. For the Support Vector Machine (SVM) classifier, the optimal hyperparameter was $C=10$.

Subsequently, with these tuned hyperparameters, we trained the three classifiers using 75\% of the available data, corresponding to 15,769 pixels. The classifiers' performance was evaluated on the test data, constituting 25\% of the dataset and comprising 5,256 pixels.
In this section, we present the results of our analysis, focusing on the overall accuracy and kappa values achieved by the three classifiers: KNN, RF, and SVM. We assess the performance of these classifiers under two dimensionality reduction techniques: PCA and LDA).

To evaluate the classifiers' performance, we employed standard confusion matrices, which allowed us to calculate classification accuracies across different classes. The results for overall accuracy and kappa values are summarized in Table III.Moreover, F-score is computed from user and producer accuracies. The predicted overall accuracies and F-score values for all datasets, methods and classes as described in Table IV.

\par
Whether it is LDA or PCA, the performance metrics, especially overall accuracy (OA) and the kappa coefficient (Ka), are promising for all three algorithms. We can observe that RF outperforms the other two with an OA of 95.5\% and a Ka of 92.6\%. It is closely followed by  KNN and SVM, which are nearly equal. KNN achieves an OA of 94.1\% and a Ka of 90.4\%, while SVM shows an OA of 93.8\% and a Ka of 89.6\%.
\par

\section{CONCLUSION}

In the realm of hyperspectral image analysis, the application of dimensionality reduction techniques has emerged as a powerful ally. This approach not only streamlines the processing of voluminous data but also mitigates the risk of computational errors. In this study, we explored the potential of LDA and PCA in enhancing the performance of various classifiers on hyperspectral data.

Our results revealed that both LDA and PCA have made impressive strides in optimizing the accuracy and efficacy of classifiers. Remarkably, PCA, with its relatively simpler method of computing principal eigenvectors, delivered outstanding outcomes, albeit with slight differences in the evaluated metrics compared to LDA.

It's worth noting that our chosen classifiers are inherently non-linear. However, when dealing with linear classifiers, LDA emerges as the more fitting choice due to its mathematical underpinnings. The subtleties of these outcomes underscore the importance of selecting the right dimensionality reduction technique, depending on the nature of the problem and the classifier being employed.

Random Forest consistently achieves high classification accuracy with both PCA and LDA dimensionality reductions, making it a robust choice for hyperspectral image classification. K-Nearest Neighbors also performs well, especially in combination with PCA. Support Vector Machines, while effective, exhibit slightly lower performance. These results underscore the importance of the judicious choice of the classifier and the potential of PCA and LDA to enhance classification accuracy in hyperspectral data

In closing, our study underscores the indispensable role of dimensionality reduction in enhancing the efficiency and accuracy of multispectral image analysis. Whether it's the streamlined simplicity of PCA or the mathematical rigor of LDA, these techniques equip us with invaluable tools to navigate the complexities of high-dimensional data. As we venture further into the realm of remote sensing and image analysis, it is clear that dimensionality reduction will continue to be a cornerstone for achieving robust, precise, and insightful results.

\printbibliography

@article{holland1995bunce,
  title={Bunce V.,(ed.) Axis, for GCSE and Standard Grade Geography. Volume 1, Number 1 (1994), 32 pp.; Volume 1, Numbers 2, 3 (1995), 32 pp. each; each magazine is accompanied by an eight-page set of teacher’s support material. Cheltenham: Stanley Thornes. Price{\pounds} 6.00 per annual magazine subscription;{\pounds} 9.99 per annual teacher’s notes subscription. ISSN 1354-800X.},
  author={Holland, Jane},
  journal={Geological Magazine},
  volume={132},
  number={6},
  pages={746--747},
  year={1995},
  publisher={Cambridge University Press}
}

@book{rees2013physical,
  title={Physical principles of remote sensing},
  author={Rees, Gareth},
  year={2013},
  publisher={Cambridge university press}
}

@book{mather2016classification,
  title={Classification methods for remotely sensed data},
  author={Mather, Paul and Tso, Brandt},
  year={2016},
  publisher={CRC press}
}

@article{abburu2015satellite,
  title={Satellite image classification methods and techniques: A review},
  author={Abburu, Sunitha and Golla, Suresh Babu},
  journal={International journal of computer applications},
  volume={119},
  number={8},
  year={2015},
  publisher={Citeseer}
}

@article{baumgardner2015220,
  title={220 band aviris hyperspectral image data set: June 12, 1992 indian pine test site 3},
  author={Baumgardner, Marion F and Biehl, Larry L and Landgrebe, David A},
  journal={Purdue University Research Repository},
  volume={10},
  number={7},
  pages={991},
  year={2015}
}

@article{sachin2015dimensionality,
  title={Dimensionality reduction and classification through PCA and LDA},
  author={Sachin, Deshmukh and others},
  journal={International journal of computer Applications},
  volume={122},
  number={17},
  year={2015},
  publisher={Citeseer}
}

@article{pang2008effective,
  title={Effective feature extraction in high-dimensional space},
  author={Pang, Yanwei and Yuan, Yuan and Li, Xuelong},
  journal={IEEE Transactions on Systems, Man, and Cybernetics, Part B (Cybernetics)},
  volume={38},
  number={6},
  pages={1652--1656},
  year={2008},
  publisher={IEEE}
}

@article{smith2002tutorial,
  title={A tutorial on principal components analysis},
  author={Smith, Lindsay I},
  year={2002}
}

@book{bishop2006pattern,
  title={Pattern recognition and machine learning},
  author={Bishop, Christopher M and Nasrabadi, Nasser M},
  volume={4},
  number={4},
  year={2006},
  publisher={Springer}
}

@book{hastie2009elements,
  title={The elements of statistical learning: data mining, inference, and prediction},
  author={Hastie, Trevor and Tibshirani, Robert and Friedman, Jerome H and Friedman, Jerome H},
  volume={2},
  year={2009},
  publisher={Springer}
}

@inproceedings{kavzoglu2020comparison,
  title={Comparison of support vector machines, random forest and decision tree methods for classification of sentinel-2A image using different band combinations},
  author={Kavzoglu, Taskin and Bilucan, Furkan and Teke, Alihan},
  booktitle={41st Asian Conference on Remote Sensing (ACRS 2020)},
  volume={41},
  pages={1--8},
  year={2020}
}

@article{breiman2001random,
  title={Random forests},
  author={Breiman, Leo},
  journal={Machine learning},
  volume={45},
  pages={5--32},
  year={2001},
  publisher={Springer}
}

@article{jung2020evaluation,
  title={Evaluation of nitrate load estimations using neural networks and canonical correlation analysis with k-fold cross-validation},
  author={Jung, Kichul and Bae, Deg-Hyo and Um, Myoung-Jin and Kim, Siyeon and Jeon, Seol and Park, Daeryong},
  journal={Sustainability},
  volume={12},
  number={1},
  pages={400},
  year={2020},
  publisher={MDPI}
}

@article{hao2010partial,
  title={Partial discharge source discrimination using a support vector machine},
  author={Hao, L and Lewin, PL},
  journal={IEEE Transactions on Dielectrics and electrical Insulation},
  volume={17},
  number={1},
  pages={189--197},
  year={2010},
  publisher={IEEE}
}

\end{document}